\documentclass[letterpaper, 10 pt, conference]{ieeeconf}  %

\IEEEoverridecommandlockouts                              %

\usepackage{graphics} %
\usepackage{amsmath} %
\usepackage{amssymb}  %
\usepackage{algorithm,algorithmic}
\usepackage{graphicx}
\usepackage{hyperref}
\usepackage{booktabs}
\usepackage{flushend}
\usepackage{xcolor}

\newcommand{\til}{\raisebox{0.5ex}{\texttildelow}}

\long\def\ignorethis#1{}
\newcommand{\myparagraph}[1]{\textbf{{#1}}}

\newcommand{\pctab}{\hspace{0.2in}}

\title{\LARGE \bf
Benchmarking Augmentation Methods for Learning Robust Navigation Agents: the Winning Entry of the 2021 iGibson Challenge
}

\author{
    Naoki Yokoyama$^{1*}$, Qian Luo$^{1*}$, Dhruv Batra$^{1, 2}$, Sehoon Ha$^{1}$ 
    \thanks{*Indicates equal contribution. $^{1}$College of Electrical Engineering and College of Computing, Georgia Institute of Technology, Atlanta, GA 30318 USA. $^{2}$DB is with Facebook AI Research (FAIR). 
    The Georgia Tech effort was supported in part by NSF, DARPA, ONR YIP, and ARO PECASE. e-mail: {\tt\scriptsize \{nyokoyama,luoqian,sehoonha,dbatra\}}@gatech.edu}
}

\begin{document}

\maketitle
\thispagestyle{empty}
\pagestyle{empty}

\begin{abstract}

Recent advances in deep reinforcement learning and scalable photorealistic simulation have led to increasingly mature embodied AI for various visual tasks, including navigation. However, while impressive progress has been made for teaching embodied agents to navigate static environments, much less progress has been made on more dynamic environments that may include moving pedestrians or movable obstacles. In this study, we aim to benchmark different augmentation techniques for improving the agent's performance in these challenging environments. We show that adding several dynamic obstacles into the scene during training confers significant improvements in test-time generalization, achieving much higher success rates than baseline agents. We find that this approach can also be combined with image augmentation methods to achieve even higher success rates. Additionally, we show that this approach is also more robust to sim-to-sim transfer than image augmentation methods. Finally, we demonstrate the effectiveness of this dynamic obstacle augmentation approach by using it to train an agent for the 2021 iGibson Challenge at CVPR, where it achieved 1\textsuperscript{st} place for Interactive Navigation.
\end{abstract}

\section{INTRODUCTION} \label{sec:introduction}
Mobile robots must be able to skillfully navigate through their environments to operate effectively in the real world. For example, home assistant robots must be able to go from room to room in order to help humans with daily chores. However, autonomous navigation is a difficult challenge that typically requires robots to understand and reason about their surroundings using visual sensors. Fortunately, several recent works using deep reinforcement learning have shown promising results by deploying robots that can successfully navigate in novel environments in the real world~\cite{srcc, yokoyama2021success}. 

One well-studied task, PointGoal Navigation~\cite{anderson2018evaluation}, challenges the agent to reach a target coordinate location in its environment. However, it typically assumes the environment is static and devoid of dynamic objects, a far-cry from real homes and offices that may contain moving humans and pets or small movable obstacles and furniture. The recent iGibson Challenge \cite{2021igibsonchallenge} at the 2021 CVPR conference was designed to encourage researchers to investigate two modified versions of PointNav that are more difficult: Interactive Navigation (InteractiveNav) and Social Navigation (SocialNav). In InteractiveNav, the agent must reach the goal while pushing aside movable obstacles that obstruct the path. For SocialNav, the agent must reach the goal while avoiding collisions with humans walking across the environment that may not yield.

These dynamic tasks impose new challenges that are absent in static navigation settings. For InteractiveNav, the agent must distinguish between moveable and unmovable objects, and learn how to push movable ones if necessary in the proper directions to form shorter paths to the goal. For SocialNav, the agent must avoid the trajectories of nearby pedestrians and reach the goal without colliding with any of them. Another factor that makes dynamic navigation tasks even more difficult is the sparsity of data available for simulating dynamic environments; while there are large datasets for static 3D environments such as Gibson \cite{Xia_2018_CVPR} and HM3D \cite{ramakrishnan2021hm3d}, datasets for dynamic environments that include individually interactable and properly arranged furniture are much sparser. For the 2021 iGibson Challenge, only eight training scenes were available.

\begin{figure}
    \centering
    \includegraphics[width=1\linewidth]{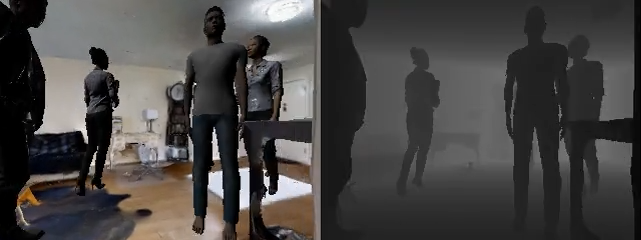}
    \caption{Dynamic obstacle augmentation adds several moving pedestrians as obstacles (visualized in RGB, left, and depth, right).
    We find that training the agent under these settings teaches it to successfully reach the goal more often in novel environments, even if navigation at test-time has no pedestrians.
    }
    \label{fig:all_agents_valid_set}
\end{figure}

To address the challenges of dynamic navigation and constraints on the availability of training scenes, we aim to provide a benchmark and analysis of augmentation techniques for visual navigation such that other researchers can leverage our findings from the iGibson Challenge. In particular, we show that simply adding several dynamic obstacles (pedestrians) to the scene during training improves the agent's test-time success rate in novel environments significantly, even for navigation tasks that \textit{do not involve dynamic obstacles in the environment}, such as PointNav and InteractiveNav. We conduct a systematic analysis in which we sweep through both the amount of training data available and the number of dynamic pedestrians used during training.

We then compare this dynamic obstacle augmentation method against 
two image augmentation methods, \textit{Crop} and \textit{Cutout}, which was shown by Laskin et al.~\cite{laskin2020reinforcement} to significantly improve test-time generalization (up to 4x performance improvement) on various benchmarks for visual tasks~\cite{laskin_srinivas2020curl,openaiprocgen}.
We show that dynamic obstacle augmentation can be combined synergistically with image augmentations to further improve performance. We also show that in contrast, combining different image augmentation methods can reduce performance gains, an observation that is also made by Laskin et al. 

Additionally, we show that dynamic obstacle augmentation is also more robust to sim-to-sim transfer. After training our agents on a dataset of 3D scans of real-world indoor scenes in the Habitat simulator \cite{szothabitat2}, we test them on a dataset of fully synthetic scenes composed by 3D artists in the iGibson simulator \cite{igibson}. We find that agents trained with dynamic obstacle augmentation are more robust to the sim-to-sim transfer, as they are able to confer higher gains in navigation performance in these synthetic scenes than agents that are trained with image augmentations.

We demonstrate the effectiveness of dynamic obstacle augmentation by participating in the 2021 iGibson Challenge at CVPR, where our agent ranked 1\textsuperscript{st} place for InteractiveNav. This feat was accomplished without using any rewards specific to InteractiveNav; only a basic PointNav reward and dynamic obstacle augmentation was used. Despite this, our approach achieved a 4\% (absolute percentage) higher success rate than the 2\textsuperscript{nd} place team.

\section{RELATED WORK}
\label{sec:related_work}

\subsection{PointGoal Navigation}
PointGoal Navigation (PointNav), as formally defined by Anderson et al. in \cite{anderson2018evaluation}, is a task in which the robot must navigate from a start position to a goal coordinate in the environment. This is a challenging task to perform in a novel scene containing obstacles, especially when the agent only has access to an egocentric camera and an egomotion sensor. Despite these constraints, Wijmans et al.~\cite{wijmans2019dd} proved that a near-perfect ($>$97\% success) PointNav agent could be trained in simulation with deep reinforcement learning. However, this work leveraged a large amount of training scenes; in this work, we focus on how augmentation methods can improve performance in scenarios in which only a small amount of scenes are available, and show that training an agent to reach the same levels of success in such scenarios is not yet solved.

\subsection{Navigation in Dynamic Environments}

Though there has been a large amount of work on autonomous robot navigation, a very small subset of this research has focused on navigation requiring the robot to interact directly with dynamic elements in the environment. The recent work by Xia et al. \cite{igibson} introduced a new task, InteractiveNav, in which the robot interacts with displaceable objects in order to reach the goal. This does not require the robot to use a manipulator; the robot can simply push past the objects using its base as it moves. InteractiveNav evaluates how well the robot can balance taking the shortest path to the goal and how much it disrupts the obstacles present in the environment. Their work showed that it is possible to teach an agent to complete this task with deep reinforcement learning, though they supply the agent with the next ten waypoints to the goal position as observations at each step. In this work, we seek to investigate performance for this task using only egocentric vision and an egomotion sensor and how it can be improved using augmentation methods.

SocialNav is another task that requires the robot to maneuver around dynamic elements in the environment, in which the robot must reach the goal without colliding with nearby moving pedestrians. Similar to our work, several recent works have used deep reinforcement learning to avoid colliding with pedestrians, such as SA-CADRL \cite{Chen2017SociallyAM} and SARL \cite{chen2019crowd}, which use the positions of the pedestrians as observations to their policy, and the work by P\'erez-D’Arpino et al. \cite{socialnav2021} which uses a motion planner that has access to a map of the environment and LiDAR data. In contrast to these works, we aim to complete the task of SocialNav without a motion planner or LiDAR, and investigate how performance can be improved using augmentation methods for visual deep reinforcement learning. We believe this setup is better suited for deployment in novel environments, where accurate localization of pedestrian positions or providing a map to a motion planner may be difficult or infeasible.

\subsection{Enhancing visual navigation performance}

Several recent works have studied techniques to improve generalization to novel environments for visual navigation. Ye et al. \cite{ye2020auxiliary} build upon the work by Wijman et al. \cite{wijmans2019dd} through the use of self-supervised auxiliary tasks, which proved to significantly improve sample efficiency. Sax et al.~\cite{sax2018mid} incorporate visual priors about the world to confer performance gains over training end-to-end from scratch without priors. Chaplot et al. \cite{chaplot2020learning, chaplot2020object} improve sample efficiency compared to typical end-to-end agents by having the agent infer information about mapping and localization, which can be supervised using privileged information from the simulator.

Unlike the works mentioned above, we study methods that yield significant improvement for visual navigation without the use of additional auxiliary losses, extra network heads, or additional sensory information (e.g., local occupancy map). In this work, we focus on augmentation methods that can be applied to virtually any visual deep reinforcement learning approach to improve test-time generalization. We study how these methods improve navigation both when used on their own and when combined with each other.

\section{METHOD} \label{sec:method}

\subsection{Augmentation Methods for Visual Navigation} \label{sec:augmentation}

Training an effective visual navigation agent requires a large dataset ($>$70 3D apartment scans) to avoid overfitting \cite{wijmans2019dd}. However, it can be difficult to have access to a diverse set of interactive training environments, especially for dynamic environments in which each object is individually interactable and must be arranged within the environment in a realistic manner, such as the synthetic scenes in the iGibson dataset, which contains only 15 scenes. For the 2021 iGibson Challenge, only eight training scenes were available. In such data-constrained scenarios, augmentation methods can be vital in improving the capability of the agent to generalize to novel environments at test-time.

\begin{figure}
    \vspace{0.05in}
    \centering
    \includegraphics[width=0.80\linewidth]{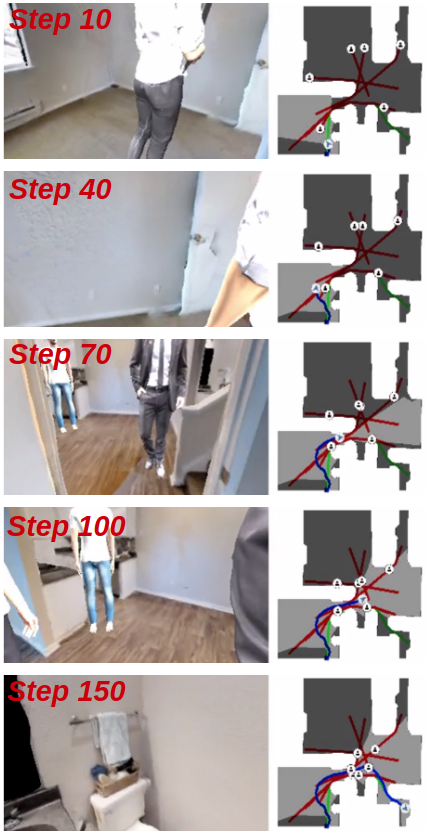}
    \caption{
Five frames from an episode of training with six dynamic pedestrians. Red denotes the pedestrian trajectories, while blue denotes the agent's trajectory. In this episode, the agent arrives at the destination without any collisions with the six pedestrians.
}
    \label{fig:nav_episode}
\end{figure}

\myparagraph{Dynamic Obstacle Augmentation.} To improve the test-time generalization of the agent for visual navigation, we introduce several moving pedestrians into the environment (illustrated in Figures~\ref{fig:all_agents_valid_set} and \ref{fig:nav_episode}). This approach aims to prevent the agent from learning to overfit to the environments it sees during training; with a large number of moving visual distractors present in the scene, it is more difficult for the agent to memorize the layout of the training environment using observations from its camera. Additionally, unlike image augmentation methods that simply perturb the agent's visual inputs, this method forces the agent to learn a larger variety of paths for a given episode's pair of start-goal positions. Because the paths that the pedestrians take are newly generated each time the environment is reset, the agent is not able to always take the same path between the same pair of start-goal positions without colliding into a pedestrian.

During training, these moving pedestrians are treated as dynamic obstacles that the agent is not allowed to collide with. If a collision between the agent and a pedestrian occurs, the episode is simply terminated. Between 3 to 12 pedestrians represented by a 3D model of a standing human from the iGibson Challenge \cite{2021igibsonchallenge} is used, which we have found to be sufficiently large (for visual occlusion) while having a small enough radius for the agent to maneuver around in most situations. At the start of an episode, for each pedestrian, two navigable points at least 3 meters apart are randomly sampled from the environment. Throughout the episode, each pedestrian moves back and forth along the shortest obstacle-free path connecting these points. It has the same maximum linear and angular speed as the agent, but its speed is randomly decreased from the maximum value by up to 10\% on a per-episode basis, such that different pedestrians within the same episode can be moving at different velocities.

\begin{figure}
    \vspace{0.1in}
    \centering
    \includegraphics[width=\linewidth]{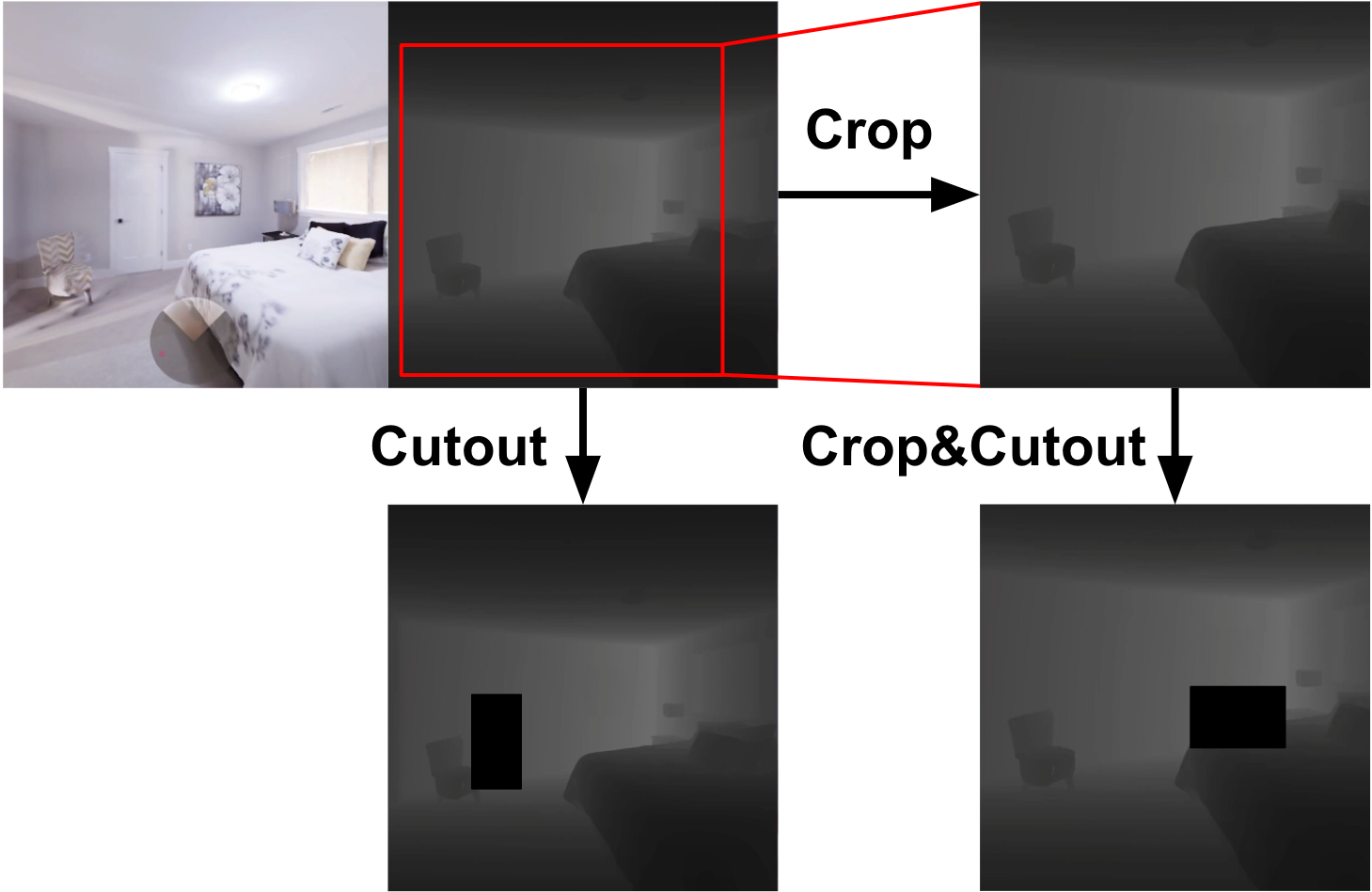}     
    \caption{
We test two image augmentations and their combination: \emph{Crop} which extracts a slightly smaller random patch, \emph{Cutout} which inserts a random black patch, and \emph{Crop\&Cutout}, which performs both.
    }
    \label{fig:image_aug}
\end{figure}

\myparagraph{Comparison with Image-based Data Augmentation.}
Image augmentation is one of the most popular approaches to improving sample efficiency for vision-based learning. Image augmentation aims to inject priors of invariance using random perturbations, such as cropping, erasing, or rotation. 
In particular, Laskin et al. \cite{laskin2020reinforcement} showed that the \emph{Crop} and \emph{Cutout} image augmentations (see Figure~\ref{fig:image_aug}) are among the most effective for visual deep reinforcement learning performance improvement. \emph{Crop} extracts a random patch from the original frame; similar to Laskin et al., we crop the frame at a random location such that the resulting image is 8\% shorter in height and width (i.e., it maintains the original aspect ratio). \emph{Cutout}, detailed by Zhong et al. \cite{zhong2020random}, inserts a black rectangle of a random shape, aspect ratio, and location into the original frame. We use the same parameters as Zhong et al., in which the rectangle's aspect ratio can be between $[0.3, 3.33]$, and its scale relative to the original frame can be between $[0.02, 0.33]$. In addition, we investigate \emph{Crop\&Cutout}, which combines \emph{Crop} and \emph{Cutout} by applying them sequentially. 

\subsection{Problem Formulation and Learning Method}

\myparagraph{Observation and Action Spaces.}
At each step, the policy observes an egocentric (first-person) depth image and its relative distance and heading to the goal point as input. As the goal point is specified relative to the agent's initial position, and the egomotion sensor indicates the agent's relative distance and heading from its initial position, they can be used to calculate the agent's current relative distance and heading to the goal. We choose to use only depth vision rather than depth and RGB vision together, as Wijmans et al.~\cite{wijmans2019dd} has shown that including RGB can hurt performance relative to using depth alone. The policy outputs a two-dimensional diagonal Gaussian action distribution, from which a pair of actions are sampled (linear and angular velocity). The maximum speeds are 0.5 m/s and 90\textdegree/s, and the policy is polled at 10 Hz. If the magnitude of the velocities are below a certain threshold (10\% of their maximum values for our experiments), it is perceived as the agent invoking a \textit{stop} action, which ends the episode. 

\myparagraph{Reward Function.}
The reward function used for all our experiments are the same regardless of what augmentation method is used, if any at all:
\begin{align*}
    r(a_t,s_t)\ =\ -\Delta_{d}-w_1-w_2(I_{back}+I_{col}) + w_3 I_{succ},
\end{align*}
where $-\Delta_{d}$ is the change in geodesic distance to the goal since the previous state, and $I_{back}$, $I_{col}$, and $I_{succ}$ are binary flags indicating whether the agent has moved backwards, collided with the environment, or terminated the episode successfully (see Success Criteria in Subsection \ref{sec:success}), respectively. $w_{1}$ serves as a constant slack penalty incurred at each step to encourage the agent to minimize episode completion time. $w_{1}$, $w_{2}$, and $w_{3}$ are set to 0.002, 0.02, and 10.0. Backward motion is penalized because we have observed that it often leads to poor visual navigation performance. However, we do not prevent it altogether, as backward movement can be helpful for avoiding an incoming obstacle that suddenly appears into the robot's view.

\myparagraph{Success Criteria.} \label{sec:success}
For all tasks, episodes are considered successful only if the agent both (1) invokes a \textit{stop} action that terminates the episode, and (2) the agent is within $0.2$~m of the goal point. Episodes are terminated and deemed unsuccessful after $500$ steps, or if the agent collides with a dynamic pedestrian either for SocialNav or when training with dynamic obstacle augmentation. A collision with a pedestrian is deemed to have occurred if the agent is within 0.3 m of it.

\begin{figure}[h]
    \centering
    \vskip5pt
    \includegraphics[width=0.48\textwidth]{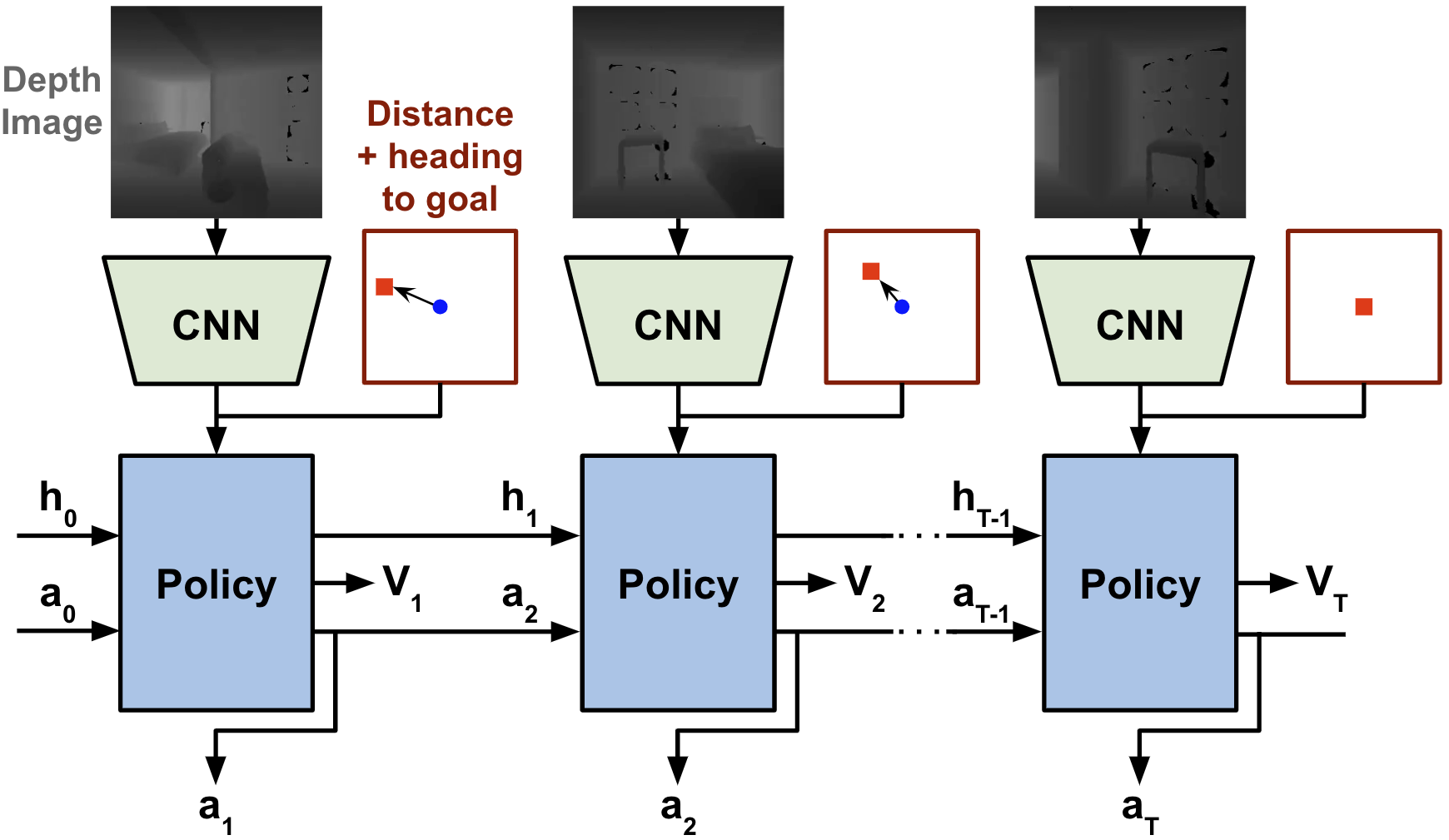}
    \caption{
The neural net is constituted by a convolutional visual encoder and a recurrent LSTM-based policy. An egocentric depth image, the agent's current distance/heading to the goal point, and the previous action is used as input. The policy outputs a Gaussian action distribution to sample linear and angular velocities, as well as an estimate of the value function for use in reinforcement learning.
    }
    \label{fig:model}
\end{figure}

\myparagraph{Network Architecture and Training.}
Our network architecture has two main components; a visual encoder and a recurrent policy. The visual encoder is a ResNet-18 \cite{resnet} convolutional neural net which extracts visual features from the depth image. The policy is a 2-layer LSTM \cite{hochreiter1997long} with a 512-dimensional hidden state. The policy receives the visual features, the relative distance and heading to the goal, the previous action, and its previous LSTM hidden state. In addition to a head that outputs actions, the policy has a critic head that outputs an estimate of the state's value, which is used for Proximal Policy Optimization (PPO) reinforcement learning as described by Schulman et al. in \cite{schulman2017proximal}. We use Decentralized Distributed PPO by Wijmans et al. \cite{wijmans2019dd} to train our agents using parallelization, and use the same learning hyperparameters. We train 8 workers per GPU, using 8 GPUs for a total of 64 workers collecting experience in parallel. Each agent is trained for 500M steps (\til280 GPU hours, or \til35 hours wall-clock).

\section{EXPERIMENTAL SETUP} \label{sec:experimental_setup}

\begin{figure*}[t]
    \centering
    \vskip5pt
    \includegraphics[width=0.99\textwidth]{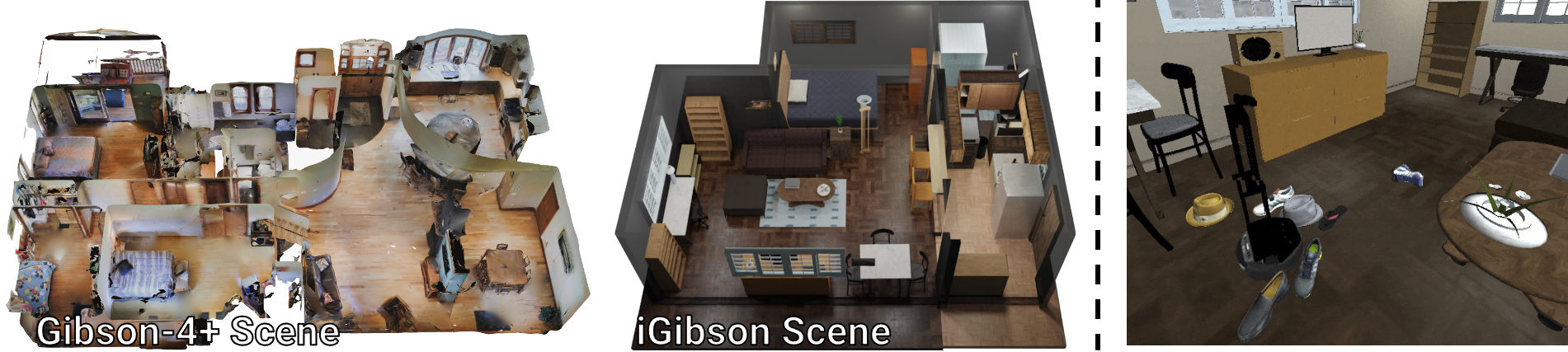}
    \caption{
\textit{Left:} For our experiments, we train and evaluate our agents in 3D scans of real-world apartments from the Gibson-4+ dataset. We use fully-synthetic, hand-designed scenes from the iGibson dataset to evaluate sim-to-sim transfer and performance in InteractiveNav, as well as for training for the iGibson Challenge.
\textit{Right:} InteractiveNav challenges the LoCoBot agent to reach the goal in the presence of movable obstacles on the ground.
    }
    \label{fig:interactivenav}
\end{figure*}

\subsection{Simulator and dataset}\label{dataset}

All of our agents are trained within Habitat \cite{habitat19iccv}, a high-performance photorealistic 3D simulator that provides the virtual robot with first-person visual data. Habitat can reach speeds of up to 3,000 frames per second per simulator instance, while the iGibson simulator used in the iGibson Challenge runs at approximately 100 frames per second \cite{szothabitat2}. Habitat also readily supports parallelization of simulator instances and gradient updates across multiple GPUs. Its speed and scalability allow us to train many agents very quickly and save a significant amount of time.

In this work, we use both the Gibson-4+ and iGibson datasets (depicted in Figure \ref{fig:interactivenav}). We use the Gibson-4+ dataset \cite{habitat19iccv} for training and evaluation, which contains 86 high quality 3D scans of real-world indoor spaces curated manually from the larger Gibson dataset \cite{Xia_2018_CVPR}. All scenes in the Gibson-4+ dataset have been rated 4 or above on a quality scale from 0 to 5 by a human, and are free of significant reconstruction artifacts such as holes in texture or cracks in floor surfaces. The training split contains 3.6M episodes distributed across the 72 scenes (50k per scene), while the validation split contains 994 episodes across 14 scenes (71 per scene). In order to select checkpoints and evaluate our agents, we randomly select (without replacement) 7 out of the 14 scenes in the validation set to form a val$_1$ and val$_2$ set, such that each set contains 497 episodes.

For the 2021 iGibson Challenge, instead of the Gibson-4+ dataset we train on the iGibson dataset of synthetic scenes (8 scenes total) imported into Habitat. We also use the iGibson dataset to evaluate agents trained on the Gibson-4+ dataset for the task of sim-to-sim transfer on InteractiveNav using the iGibson simulator.

In all cases, the agent's height and radius are modeled after the LoCoBot platform (0.8 m and 0.18 m) \cite{locobot}. The LoCoBot is depicted on the right of Figure \ref{fig:interactivenav}.

\subsection{Evaluation Details}

During evaluation, the agent's behavior was set to be deterministic for reproducibility, by sampling only the means from the policy's output Gaussian action distribution. We trained each agent configuration three times using different seeds. For each seed, the checkpoint that yielded the highest success rate on val$_1$ was selected, and reported evaluation metrics are based on that checkpoint's performance on val$_2$. There are no pedestrians present during PointNav evaluation, and three pedestrians for SocialNav evaluation. For InteractiveNav, objects from the YCB dataset \cite{calli2015ycb} are placed on the shortest obstacle-free path from the start to the goal spaced 0.5 m apart.

\subsection{Evaluation Metrics}

We use success rate as our primary evaluation metric. Apart from success, to evaluate InteractiveNav performance in iGibson, we use Success weighted by inverse Path Length (SPL)~\cite{anderson2018evaluation}, as well as Effort Efficiency and Interactive Navigation Score (INS)~\cite{xia2020interactive}. SPL and Effort Efficiency are both scores between 0 and 1, with 1 being the best. SPL measures how short the agent's path is relative to the length of the shortest path between the start and goal. Effort Efficiency penalizes displacement and forces that are applied to objects in the environment by the robot. INS is simply the average of SPL and Effort Efficiency.

\section{RESULTS} \label{results}
This section evaluates the effects of different augmentation methods for learning various visual navigation tasks. We design the experiments to answer the following questions: 
\begin{enumerate}
    \item How does dynamic obstacle augmentation affect navigation performance across a wide range of training data availability?
    \item What are the effects of combining dynamic obstacle augmentation with image augmentation techniques?
    \item How well does each augmentation method handle the gap from the Gibson 4+ to the iGibson dataset?
\end{enumerate}
We present all metrics multiplied by 100 for readability.
\setlength{\tabcolsep}{3pt}
\begin{table}[h]
\vspace{0.2cm}
\centering
\caption{
    \textbf{PointNav (0 ppl)} Success Rates
}
\label{tab:dynamic_pointnav_table}
\resizebox{\columnwidth}{!}{
    \begin{tabular}{cccccc}
        \toprule
        \# of & \multicolumn{5}{c}{\# of dynamic pedestrians during training} \\
        train & (baseline) & \multicolumn{4}{c}{}  \\

        scenes & 0 ppl & 3 ppl & 6 ppl & 12 ppl & 18 ppl \\
        \cmidrule {2-6}
        1 & 74.93\tiny{$\pm$}1.91 & 73.90\tiny{$\pm$}3.26 & \textbf{78.64\tiny{$\pm$}1.76} & 76.53\tiny{$\pm$}2.65 & 78.12\tiny{$\pm$}0.57 \\
        2 & 81.41\tiny{$\pm$}2.03 & 80.85\tiny{$\pm$}1.30 & 81.22\tiny{$\pm$}0.89 & 81.69\tiny{$\pm$}1.17 & \textbf{82.86\tiny{$\pm$}1.26} \\
        4 & 82.39\tiny{$\pm$}0.87 & \textbf{86.15\tiny{$\pm$}0.57} & 84.37\tiny{$\pm$}1.79 & 84.32\tiny{$\pm$}2.07 & 85.77\tiny{$\pm$}0.61 \\
        8 & 87.89\tiny{$\pm$}1.31 & \textbf{91.41\tiny{$\pm$}0.80} & 88.97\tiny{$\pm$}1.65 & 90.99\tiny{$\pm$}1.52 & 89.44\tiny{$\pm$}0.34 \\
        16 & 91.60\tiny{$\pm$}1.68 & 90.99\tiny{$\pm$}0.98 & 91.69\tiny{$\pm$}0.53 & \textbf{92.07\tiny{$\pm$}0.81} & 91.69\tiny{$\pm$}0.70 \\
        32 & 90.80\tiny{$\pm$}0.78 & 92.68\tiny{$\pm$}0.40 & \textbf{93.90\tiny{$\pm$}0.65} & 92.68\tiny{$\pm$}1.33 & 92.39\tiny{$\pm$}0.69 \\
        72 & 94.84\tiny{$\pm$}1.50 & \textbf{95.26\tiny{$\pm$}0.46} & 93.43\tiny{$\pm$}1.16 & 94.93\tiny{$\pm$}0.83 & 94.08\tiny{$\pm$}0.61 \\
        \bottomrule
    \end{tabular}
}
\end{table}

\setlength{\tabcolsep}{3pt}
\begin{table}[h]
\centering
\caption{
\textbf{SocialNav (3 ppl)} Success Rates
}
\label{tab:dynamic_socialnav_table}
\resizebox{0.9\columnwidth}{!}{
    \begin{tabular}{ccccc}
        \toprule
        \# of & \multicolumn{4}{c}{\# of dynamic pedestrians during training} \\
        train & (baseline) & \multicolumn{3}{c}{}  \\

        scenes & 3 ppl & 6 ppl & 12 ppl & 18 ppl \\
        \cmidrule {2-5}
        1 & 65.26\tiny{$\pm$}2.47 & \textbf{68.78\tiny{$\pm$}0.35} & 65.40\tiny{$\pm$}3.77 & 68.40\tiny{$\pm$}2.26 \\
        2 & 68.97\tiny{$\pm$}0.96 & 69.20\tiny{$\pm$}2.92 & \textbf{72.68\tiny{$\pm$}2.03} & 71.31\tiny{$\pm$}1.66 \\
        4 & 74.60\tiny{$\pm$}1.45 & \textbf{77.32\tiny{$\pm$}0.64} & 76.10\tiny{$\pm$}0.24 & 75.77\tiny{$\pm$}1.52 \\
        8 & 80.38\tiny{$\pm$}1.27 & \textbf{80.89\tiny{$\pm$}1.04} & 79.91\tiny{$\pm$}3.37 & 78.64\tiny{$\pm$}1.04 \\
        16 & 83.00\tiny{$\pm$}1.37 & \textbf{83.71\tiny{$\pm$}0.59} & 82.96\tiny{$\pm$}1.61 & 80.47\tiny{$\pm$}0.63 \\
        32 & 84.04\tiny{$\pm$}1.09 & \textbf{84.41\tiny{$\pm$}0.74} & 83.99\tiny{$\pm$}1.29 & 81.27\tiny{$\pm$}0.87 \\
        72 & 85.26\tiny{$\pm$}1.27 & 84.79\tiny{$\pm$}1.33 & \textbf{86.43\tiny{$\pm$}1.75} & 83.10\tiny{$\pm$}2.93 \\
        \bottomrule
    \end{tabular}
}
\end{table}

\subsection{Performance with Dynamic Obstacle Augmentation}

We evaluated the performances of many training configurations, in which both the amount of available training scenes and number of dynamic pedestrians used during training are varied. Table~\ref{tab:dynamic_pointnav_table} represents the success rates for PointNav, while Table~\ref{tab:dynamic_socialnav_table} shows the success rates for SocialNav. We did not evaluate agents trained with zero pedestrians for SocialNav due to their inability to adapt to the presence of dynamic obstacles at test-time, which led to poor performance. 

\myparagraph{Finding 1: Dynamic obstacle augmentation improves visual navigation performance, even when the environment is static.}
Our results indicate that dynamic obstacle augmentation improves performance for both visual navigation tasks, as the best performance is achieved when dynamic pedestrians were added in the training environments. An interesting observation is that dynamic obstacle augmentation improves test-time generalization even for PointNav, where the environment is completely static and there are no dynamic obstacles at test-time. 

\myparagraph{Finding 2: Dynamic obstacle augmentation is more effective with a small number of training scenes.}
Both tables also show that the performance gains provided by dynamic obstacle augmentation are more substantial when only a small number (8 or less) of training scenes are available. As the number of available training scenes increases, the gap between the baseline agents and those trained with added dynamic pedestrians decreases. For instance, the difference between the baseline and the best agents is $3.5$\% with eight training scenes but the difference is only $0.42$\% with $72$ training scenes in Table~\ref{tab:dynamic_pointnav_table}. This trend indicates that the limited number of training scenes causes the agent to overfit, but can be mitigated by dynamic obstacle augmentation.

\subsection{Comparison with Image Augmentation Methods}
\label{sec:compare_aug}
\begin{figure}[h]
    \centering
    \vspace{-8pt}
    \includegraphics[width=0.48\textwidth]{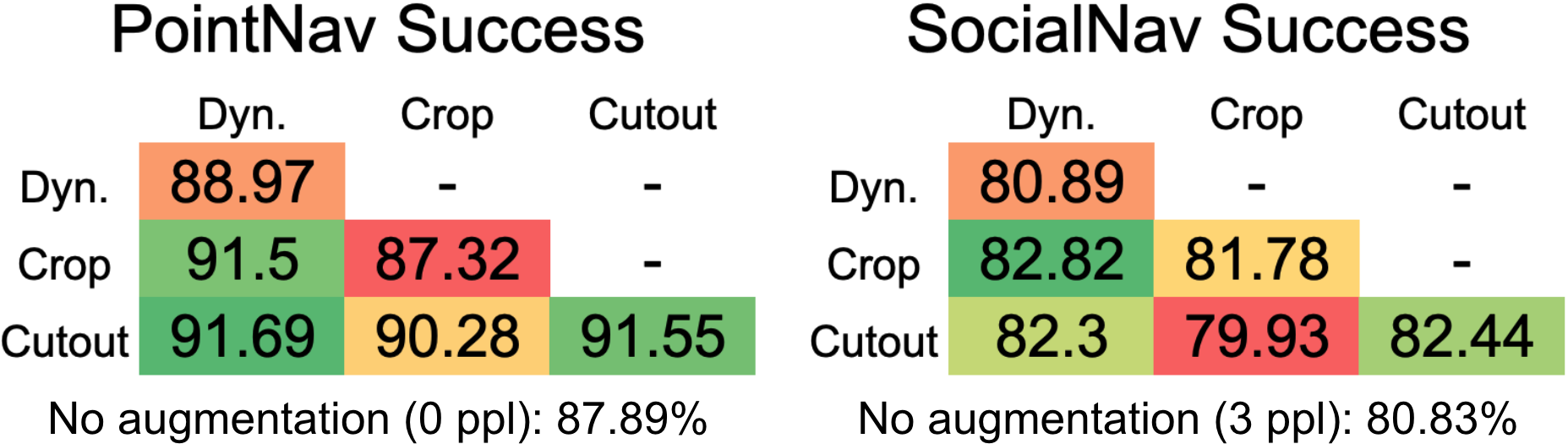}
    \caption{
Heatmaps comparing average success rates over three random seeds when trained with various augmentation methods. \emph{Crop\&Cutout} is less successful than \emph{Cutout} alone, while adding \textit{Dyn} leads \textit{Crop} to attain the highest success rates for both tasks and does not decrease the performance of \textit{Cutout}.
    }
    \label{fig:data_aug_heatmap}
\end{figure}

In this section, we compare our technique against image augmentation techniques, \emph{Crop} and \emph{Cutout}, as defined in Section~\ref{sec:augmentation}. To narrow the scope of our investigation and evaluate how well each technique mitigates overfitting, we limit experiments to agents trained with only eight scenes. When dynamic obstacle augmentation is used, six pedestrians are added during training. Otherwise agents were trained with zero pedestrians for PointNav, or three pedestrians for SocialNav. 

\myparagraph{Finding 3: Image augmentation can improve test-time success rate as well, yet combining them can decrease performance gains.}
We find that \emph{Crop} and \emph{Cutout} can also improve test-time success when each augmentation is used alone as shown in Figure~\ref{fig:data_aug_heatmap}, with the exception of \emph{Crop} for PointNav. We find that using \emph{Cutout} alone outperforms dynamic obstacle augmentation by a slight margin for both tasks. We also evaluate the effects of combining the image augmentation techniques. When training with deep reinforcement learning, applying different data augmentation techniques in conjunction does not necessarily improve performance over a single isolated technique. Laskin et al.~\cite{laskin2020reinforcement} showed that the best results on the DeepMind Control Suite~\cite{laskin_srinivas2020curl} are achieved when only \emph{Crop} is used, and performance would only drop when combined with other image augmentation techniques. Similarly, we observe that combining \emph{Cutout} and \emph{Crop} (bottom-middle cells) only leads to worse performance than simply using \emph{Cutout} alone.

\myparagraph{Finding 4: Combining dynamic obstacle and image augmentation techniques can further increase success rates.}
However, as shown in Figure \ref{fig:data_aug_heatmap}, when dynamic obstacle augmentation is combined with either the \emph{Crop} or \emph{Cutout} image augmentations, we find that the success rate can improve even further than using either method alone. While combining image augmentation methods hurts performance, combining \emph{Dyn} with \emph{Crop} (middle-left cells) yields a higher success rate than using either method alone. This combination achieves the highest success rate for both tasks, tying with \emph{Cutout} for PointNav and surpassing it for SocialNav. Combining \emph{Cutout} with \emph{Dyn} (bottom-left cells) does not affect performance over using \emph{Cutout} alone, in contrast to the performance decrease that occurs when \emph{Crop} is added to \emph{Cutout}.

\subsection{Sim-to-Sim Transfer for iGibson InteractiveNav}
\label{sec:interactivenav}
\begin{table} 
\vspace{0.21cm}
\begin{center}
\caption{\label{tab:results1}Generalization to iGibson InteractiveNav}   
\label{tab:interactive_nav_table}
\begin{tabular}{lllll}    
\toprule  & Success & SPL & EffortEfficiency & INS\\    \midrule  
 No Augmentation & 51.00\tiny{$\pm$}8.03 & 13.10\tiny{$\pm$}2.47 & 92.70\tiny{$\pm$}1.28 & 52.90\tiny{$\pm$}1.85\\  
 Dynamic & \textbf{61.99\tiny{$\pm$}3.30} & \textbf{15.58\tiny{$\pm$}0.91} & \textbf{94.15\tiny{$\pm$}1.09} & \textbf{54.86\tiny{$\pm$}0.74}\\
 Dynamic+crop & 61.20\tiny{$\pm$}6.80 & 15.51\tiny{$\pm$}2.11 & 92.83\tiny{$\pm$}1.30 & 54.17\tiny{$\pm$}1.70 \\
 Dynamic+cutout & 60.54\tiny{$\pm$}3.87 & 14.77\tiny{$\pm$}0.62 & 93.04\tiny{$\pm$}1.19 & 53.77\tiny{$\pm$}0.65 \\  
 Crop & 56.11\tiny{$\pm$}1.32 & 14.02\tiny{$\pm$}0.52 & 92.64\tiny{$\pm$}0.86 & 53.40\tiny{$\pm$}0.61 \\ 
 Cutout & 54.96\tiny{$\pm$}2.35 & 13.78\tiny{$\pm$}1.01 & 92.83\tiny{$\pm$}1.30 & 53.04\tiny{$\pm$}1.13\\
 Crop+Cutout & 52.40\tiny{$\pm$}3.56 & 13.13\tiny{$\pm$}1.99 & 92.20\tiny{$\pm$}1.45 & 52.66\tiny{$\pm$}1.66\\
 \bottomrule   
\end{tabular}  

\end{center}
\end{table}
Though we can evaluate our agents for PointNav and SocialNav in Habitat, emulating the InteractiveNav task in Habitat requires a large amount of engineering and remained out of our reach for this work, despite prolonged effort. Instead, we evaluate the agents trained in Habitat directly in iGibson for InteractiveNav in a zero-shot manner (no adaptation). This is the same approach we used to achieve 1\textsuperscript{st} place in the iGibson Challenge.

As in the previous subsection, all agents were trained with eight Gibson-4+ scenes, and those trained with dynamic obstacle augmentation were trained with six pedestrians (no pedestrians otherwise). We evaluate our agents in the eight iGibson scenes from the training set for the InteractiveNav task; currently, the evaluation set from the iGibson Challenge is not public. 

\myparagraph{Finding 5: Dynamic obstacle augumentation is more robust to sim-to-sim transfer from Gibson-4+ to iGibson scenes.}
As shown in Table~\ref{tab:interactive_nav_table}, while all augmentation methods improve performance over using no augmentation at all, dynamic obstacle augmentation did so by the greatest amount (\til11\% more success). Notably, in contrast to the results in the previous subsection, the results yielded by the image augmentation methods are much less competitive against dynamic obstacle augmentation. In fact, when using \textit{Crop} or \textit{Cutout} in conjunction with our method, performance is actually slightly reduced. Whereas \textit{Cutout} by itself conferred one of the highest gains in performance in the previous subsection, it now provides the least gains over the baseline, beating only \textit{Crop\&Cutout}.

We attribute this to the visual gap between the iGibson scenes and the Gibson-4+ scenes that our agents were trained on; Gibson-4+ scenes are collected from the real world using a 3D camera and may contain some irregularities or artifacts, whereas iGibson scenes are completely synthetic, comprised wholly by 3D computer-aided design (CAD) files made by artists, and thus contain no artifacts and have much more geometric surfaces and edges (see Figures \ref{fig:all_agents_valid_set} and \ref{fig:interactivenav} for comparison). 

Because image augmentation methods encourage the agent to infer the removed information from the remaining context available in the visual input, agents trained with these methods are more sensitive to visual changes in that context. In contrast, dynamic obstacle augmentation maintains a substantial performance gain. We conjecture that it allows agents to learn a better spatial understanding of their surroundings by dodging obstacles and recalculating new paths, as opposed to simply teaching the agents to use existing visual context to recover from visual perturbations.

\begin{table} 
\vspace{0.21cm}
    \begin{center}
        \caption{\label{tab:eval_ai}2021 iGibson Challenge Success Rates}
        \resizebox{0.75\columnwidth}{!}{
            \begin{tabular}{lcc}    
\toprule  
Team    & InteractiveNav                     & SocialNav                     \\
\midrule 
Ours    & \textbf{31\%}                      & 26\%                          \\
LPAIS   & 27\%                               & 26\%                          \\
NICSEFC & 26\%                               & \textbf{28\%}                 \\
LPACSI  & -                                  & \textbf{28\%}                 \\
SYSU    & 26\%                               & -                             \\
Fractal & -                                  & 25\%                          \\
             \bottomrule   
            \end{tabular}  
        }
    \end{center}
\end{table}

\subsection{Results on 2021 iGibson Challenge}
For the agent we submitted to the 2021 iGibson Challenge, \textit{we train on the 8 provided iGibson dataset scenes, not on Gibson-4+ scenes} using the Habitat simulator. We submitted the same agent to both the InteractiveNav and SocialNav tracks. The results of the top five teams for each track are shown in Table \ref{tab:eval_ai}. Dynamic obstacle augmentation was the only augmentation used, with 12 pedestrians being present during training. For InteractiveNav, we achieved 4\% more success than the 2\textsuperscript{nd} place team, who used image augmentation and a reward function very similar to ours, except with added terms designed to encourage the robot to seek alternative paths when blocked \cite{youtube_2021}. Additionally, their agent is trained natively in iGibson, precluding any issues that may arise from sim-to-sim transfer from Habitat to iGibson (which we discuss at the end of this section). For SocialNav, we attain a success rate only 2\% lower than the 1\textsuperscript{st} place team.

Episode visualizations and metrics for causes of episode failure on the challenge scenes were not shown to participants. Upon asking the organizers, we were informed that the large majority of episodes across all entries failed due to timeout (71\% for InteractiveNav and 63\% for SocialNav). We speculate that the large difference between the success rates on the challenge scenes (around $30$\%) and the experiment results seen in subsections \ref{sec:compare_aug} and \ref{sec:interactivenav} (around $81$\% and $60$\%), in which only eight training scenes were also used (albeit from Gibson-4+, not iGibson), may be due to the size and layout of the challenge scenes and the sim-to-sim gap from Habitat to the iGibson simulator. Scenes from the iGibson dataset tend to be relatively small (see Figure \ref{fig:interactivenav}), especially compared to the real-world scans of the Gibson-4+ dataset used to train the agents in subsections \ref{sec:compare_aug} and \ref{sec:interactivenav}. The size and furniture layout of the challenge scenes may have restricted the robot to narrower paths that caused it to get caught on obstacles more often, or accidentally block a path to the goal with moveable furniture or obstacles (especially for InteractiveNav, where there is also clutter), or make passing an incoming pedestrian along a narrower path much more difficult for SocialNav. 

Additionally, we ``freeze'' the iGibson training scenes in Habitat to maintain high simulation speeds when training our agents. In other words, furniture in the scene such as chairs and lamps could not be moved by the robot in the Habitat simulator, whereas they could be moved in either InteractiveNav or SocialNav upon deployment in the iGibson simulator. Our agents avoid pushing larger objects because these objects cannot move in training, and thus the agents learn to simply try to circumvent them entirely. This may have also contributed to making navigation more difficult for the robot in the iGibson Challenge, as it may not have been able to adapt to cases in which colliding with the environment caused an obstacle to block the path to the goal.

\section{CONCLUSION AND FUTURE WORK}
In this work, we conducted a systematic analysis on augmentation methods to improve the performance of deep reinforcement learning methods for visual navigation tasks: PointNav, SocialNav, and InteractiveNav. We show that the dynamic obstacle augmentation approach detailed in this work can significantly improve test-time generalization, be synergistically combined with image augmentation methods for further improvements in success, and are more robust to sim-to-sim transfer than image augmentation methods.

There are several directions in which our work can be extended. We suspect that while using dynamic pedestrians can boost performance, overcrowding can make the agent overly conservative and result in suboptimal behaviors. In the future, we plan to introduce curriculum learning to gradually increase the number of pedestrians during training, in order to potentially increase the number of pedestrians that will maximize performance.

We also plan to investigate sim-to-real transfer of the proposed technique by deploying it to real-world with pedestrians. However, we expect that the sim-to-real gap will lead to significant drops in performance, especially because pedestrians used within simulation are single rigid bodies (i.e., limbs are not independently animated). This could be addressed by adding more realistic motions and gestures for simulated humans to reduce the sim-to-real gap.

\section{Acknowledgements}
\scriptsize{
    The Georgia Tech effort was supported in part by NSF, ONR YIPs, ARO PECASE. The views and conclusions are those of the authors and should not be interpreted as representing the U.S. Government, or any sponsor.
}

\bibliography{ref}
\bibliographystyle{IEEEtran}
\end{document}